\documentclass{article}

\usepackage{PRIMEarxiv}

\usepackage[utf8]{inputenc} 
\usepackage[T1]{fontenc}    
\usepackage{hyperref}       
\usepackage{url}            
\usepackage{booktabs}       
\usepackage{amsfonts}       
\usepackage{nicefrac}       
\usepackage{microtype}      
\usepackage{lipsum}
\usepackage{fancyhdr}       
\usepackage{graphicx}       
\graphicspath{{media/}}     
\usepackage{todonotes}
\usepackage{subfig}

\DeclareUnicodeCharacter{0307}{\.{}} 

\title{TIPTA UZMANLIK SINAVINDA (TUS) BÜYÜK DİL MODELLERİ İNSANLARDAN DAHA MI BAŞARILI?

}

\author{
  Yesim Aygul, Muge Olucoglu, Adil Alpkocak \\
  Department of Computer Engineering \\
  Izmir Bakircay University \\
  Izmir, Turkey\\
  \texttt{\{yesim.aygul, muge.olucoglu, adil.alpkocak\}@bakircay.edu.tr} \\
}

\begin{document}
\maketitle

\centerline
  {\large \bfseries \scshape Özet}

\begin{quote}
Yapay zekanın, doğal dil işleme alanındaki son gelişmeler, yapay zekanın tıp eğitimi ve değerlendirmesindeki potansiyelini ortaya çıkarmıştır. Bugün yapay zeka modelleri tıbbi soruları başarı ile yanıtlayabilmektedir. Sağlık profesyonellerine yardımcı olabilmektedir. Bu çalışma, üç farklı yapay zeka modelinin 2021 1. Dönem Tıpta Uzmanlık Sınavı'ndaki (TUS) Türkçe tıp sorularını yanıtlamadaki performanslarını değerlendirmektedir. TUS, klinik (KTBT) ve temel (TTBT) tıp bilimleri genelinde toplamda 240 sorudan oluşmaktadır. KTBT'deki sonuçlara göre Gemini'nin 82 soruyu, ChatGPT-4'ün 105 soruyu ve ChatGPT-4o'nun 117 soruyu doğru yanıtladığı sonucu elde edilmiştir. TTBT'de ise Gemini ve ChatGPT-4 93, ChatGPT-4o 107 soruyu cevap anahtarına göre doğru cevaplamıştır. ChatGPT-4o, KTBT ve TTBT'ye göre sırayla en yüksek 113 ve 106 puanı alan adaydan daha iyi performans göstermiştir. Bu çalışma, yapay zekanın tıp eğitimi ve değerlendirmedeki potansiyelinin önemini vurgulamaktadır. Gelişmiş modellerin yüksek doğruluk ve bağlamsal anlayış elde edebildiğini göstererek bunların tıp eğitimi ve değerlendirmesindeki potansiyel rolünü göstermektedir.
\end{quote}

\anahtarKelimeler{Büyük Dil Modelleri \and ChatGPT \and Gemini \and Tıpta Uzmanlık Sınavı\and Tıp Eğitimi }\vspace{1cm}

\begin{abstract}
The potential of artificial intelligence in medical education and assessment has been made evident by recent developments in natural language processing and artificial intelligence. Medical questions can now be successfully answered by artificial intelligence algorithms. It can help medical practitioners. This study evaluates the performance of three different artificial intelligence models in answering Turkish medical questions in the 2021 1st Term Medical Specialization Examination (MSE). MSE consists of a total of 240 questions across clinical (CMST) and basic (BMST) medical sciences. According to the results in CMST, it was concluded that Gemini correctly answered 82 questions, ChatGPT-4 answered 105 questions and ChatGPT-4o answered 117 questions. In BMST, Gemini and ChatGPT-4 answered 93 questions and ChatGPT-4o answered 107 questions correctly according to the answer key. ChatGPT-4o outperformed the candidate with the highest scores of 113 and 106 according to CMST and BMST respectively. This study highlights the importance of the potential of artificial intelligence in medical education and assessment. It demonstrates that advanced models can achieve high accuracy and contextual understanding, demonstrating their potential role in medical education and evaluation.
\end{abstract}

\newpage
\section{Giriş}

Türkiye'de Tıpta Uzmanlık Sınavı (TUS) olarak bilinen sınav, tıp doktorlarının uzmanlaşmak istedikleri alanlarda eğitim almak için uzmanlık eğitimine giriş amacıyla yapılan bir sıralama başarıları sınavıdır \cite{cansever2020tip}. Bu sınav, bir adayın, klinik bilgilerini ve tıbbi kavramlar bilgilerini test etmek üzere tasarlanmıştır. Bu sınavların karmaşıklığı, hazırlık ve değerlendirme için yenilikçi yaklaşımlar gerektirmektedir. 

Bu makale, TUS sorularının çözümünde gelişmiş yapay zeka modellerinden üç tanesini kullanarak sonuçlarını araştırmakta, performanslarının karşılaştırılması ve değerlendirmesi için potansiyelini sunmaktadır. Bu karşılaştırmalı çalışmada yöntem olarak, yapay zeka modellerinin her birine  bir TUS soru seti girdi olarak verilip sonuçları kaydedilmiştir. Çalışmada 2021 yılına ait 1. Dönem TUS soruları \cite{tusklinik, tustemel} ile çalışılmıştır. Bu döneme ait sınavda toplam 240 adet soru bulunmaktadır. Bu sorular anatomi, fizyoloji, patoloji ve klinik senaryolar dahil olmak üzere geniş bir tıbbi konu yelpazesini kapsamaktadır. Her modelin yanıtları doğruluklarına ve açıklama derinliklerine göre değerlendirilmiştir. Bu metadoloji, her modelin başarımının değerlendirilmesine olanak tanımaktadır.
 
ChatGPT-4, ChatGPT-4o ve Gemini gibi yapay zeka modelleridir. Bu BDM teknolojileri, zorlu konularda anında geri bildirim ve farklı bakış açıları sunarak öğrenciler için tamamlayıcı araçlar olarak hizmet edebilmektedir. Dahası, eğitimcilere yaygın kavram yanılgılarını ve öğrencilerin zorlandıkları alanları tespit etmede yardımcı olabilir ve böylece öğretim stratejilerini daha etkili bir şekilde uyarlayabilmektedir. Bununla birlikte, tıbbi kavramların ve klinik becerilerin kapsamlı bir şekilde anlaşılmasını sağlamak için yapay zekaya olan güven geleneksel öğrenme yöntemleriyle birlikte dengeli ilerlemelidir.

Bu çalışma, Gemini, ChatGPT-4 ve ChatGPT-4o gibi gelişmiş yapay zeka modellerinin tıp eğitiminde ve değerlendirme süreçlerinde nasıl kullanılabileceğini göstermektedir. Bu modellerin TUS sorularını çözme yeteneklerini kapsamlı bir şekilde ortaya koymuştur. Bu çalışma, yapay zekanın, açıklama ve doğruluk gibi yetenekleri ele alındığında tıp eğitimi ve değerlendirme süreçlerine katkı sağlayacağı sonucuna varılmıştır. Yapay zeka modellerinin tıp eğitiminde yeni uygulamalar için araştırma konuları ve çok çeşitli uygulama alanları sunabileceğini göstermiştir. 

Bu çalışmadaki temel amaç, yapay zeka modellerindeki teknolojilerinin hızlı gelişimi ve farklı yanıt verme yeteneklerinin değerlendirilmesini ortaya koymaktır. Bu çalışma, ChatGPT-4, ChatGPT-4o ve Google Gemini yapay zeka modellerinin performansını 2021 yılına ait 1. Dönem TUS'a ait 240 soru üzerinden verilen cevaplarla karşılaştırmayı amaçlamaktadır. Bu karşılaştırma, yapay zeka teknolojilerinin gelişimini ve farklılıklarını analiz etmeye olanak tanımaktadır. Bu çalışma, tıbbi eğitim ve sınav hazırlığı gibi spesifik alanlarda hangi yapay zeka modelinin daha yararlı olduğunu tespit ederek, kullanıcılar için en uygun öğrenme araçlarının seçilmesine yardımcı olmayı hedeflemektedir.

Bu çalışmanın ikinci bölümünde Büyük Dil Modellerinin sağlık alanında kullanım yöntemlerinden ve Türkiye’de yapılan Tıpta Uzmanlık Sınavı’na benzer yurt dışındaki tıp eğitiminde kullanılan sınavların yapılan yapay zeka modelleri ile çözümlerinden bahsedilecektir. Çalışmanın üçüncü bölümünde ise kullanılan yöntemler Open AI’ın geliştirdiği ChatGPT-4, ChatGPT-4o, ve Google’ın geliştirdiği Gemini modellerinden ve bu modellerin birbirleriyle kıyaslamasından bahsedilmektedir. Dördüncü bölümde ise modellerin kıyaslamalı sonuçları başarım ile birlikte bulgular ele alınmakta olup sunulmaktadır. Beşinci bölümde, tüm makale özetlenerek elde edilen başarım sonuçları sunulmaktadır.

\section{Büyük Dil Modelleri ve Tıbbi Uygulamaları}

Büyük Dil Modelleri (BDM) çeşitli alanlarda uygulamalarıyla büyük bir devrim yaratmıştır. Bu modeller, çeşitli veri kümelerinde kapsamlı eğitime dayalı olarak insan benzeri metin işleme ve üretme yetenekleriyle karakterize edilmektedir \cite{joachimiak2024artificial}. BDM'nin tarihsel gelişimi, araştırmacıların insan benzeri metinleri anlayabilen ve üretebilen sistemler oluşturmayı amaçladıkları Doğal Dil İşleme (DDİ) ve makine öğrenimi alanındaki ilk çalışmalara kadar uzanmaktadır. BDM'nin temel özellikleri arasında çok miktarda metin verisinden öğrenme, karmaşık dil kalıplarını yakalama ve tutarlı ve bağlamsal olarak ilgili metinler üretme becerileri yer almaktadır. Bu modeller tipik olarak girdi metnini işleyen ve öğrenilen kalıplara dayalı çıktı üreten çok katmanlı sinir ağlarından oluşmaktadır. Bu da BDM’leri farklı alanlardaki çeşitli uygulamalar için çok yönlü araçlar haline getirmektedir.
 
BDM’ler, büyük miktarda metinsel veriyi işlemelerini sağlayan dönüştürücü mimariler kullanılarak oluşturulmuştur. Bu modeller, çok çeşitli konuları ve kavramları kapsayan milyarlarca kelime içeren derlemler üzerinde önceden eğitilmiştir. Ön eğitim aşaması, BDM'lerin deyimsel ifadeler ve alana özgü terminolojiler de dahil olmak üzere incelikli bir dil anlayışı geliştirmesine olanak tanımaktadır. Özel veri kümeleri üzerinde yapılan ince ayarlar, belirli görevleri yerine getirme yeteneklerini daha da geliştirerek onları sağlık alanında kolaylıklar sağlamaktadır.

BDM’ler birçok farklı alanda kullanılarak, kullanıcılarına önemli katkılar sağlamaktadır. Bu gelişmiş dil modelleri, teşhis ve tedavi için yeni olanaklar sunmuştur. BDM'ler, tıbbi metinlerin analiz edilmesi, hasta sorularına yanıt verilmesi, tıbbi makalelerin özetlenmesi ve klinik karar destek sistemleri gibi birçok alanda modellenerek kullanılmaktadır.  Bu modeller, hasta-doktor etkileşimlerinde ve tıbbi eğitimde de kullanılabilirlik göstermiştir. Tıp alanında, BDM'ler klinik karar destek sistemlerinde önemli bir rol oynamaktadır \cite{ong2024advancing}. Bu sistemlerden, modellerin karmaşık hasta verilerini analiz etme ve kanıta dayalı öneriler üretmek için yararlanılmaktadır. Örneğin, BDM'ler, ayırıcı tanılar önermek için hasta semptomlarını ve tıbbi geçmişi sentezleyerek tanı sürecine yardımcı olabilmektedir. Bu işlev, birden fazla durumun birbiriyle uyumlu semptomlar sergilediği özellikle karmaşık vakalarda daha da önemlidir. Böylece klinisyenlerin daha bilinçli kararlar almasına destek olmasına yardımcı olmaktadır.
 
BDM'ler tıp eğitimi ve öğretiminde de önemli bir rol oynamaktadır. Tıp öğrencilerine ve profesyonellere karmaşık tıbbi kavramları açıklayan, soruları yanıtlayan ve simüle edilmiş hasta modelleri ile eğitim alanına hizmet vermektedirler \cite{kim2024health}. Bu simülasyon uygulama modeli, anında geri bildirim ve kişiselleştirilmiş öğrenme yolları sunarak eğitim deneyimini geliştirmektedir. Örneğin; bir BDM, hasta görüşmelerini simüle etmek için kullanılabilir ve öğrencilerin risksiz bir ortamda tanısal muhakeme ve klinik karar verme pratiği yapmalarına olanak tanımaktadır \cite{li2023meddm}. 

BDM, tıbbi araştırmalarda, hastalıkların anlaşılması, yeni tedavilerin geliştirilmesinin yanında ilaç keşif süreçlerinin hızlandırılmasındaki ilerlemelere önemli ölçüde katkıda bulunmuştur. Bu modeller, büyük miktarlardaki biyomedikal literatürü, genetik verileri ve klinik deneyleri analiz ederek gizli kalıpları ortaya çıkarmaktadır. Bu durum, ilaç etkileşimlerini tahmin edebilir kılmaktadır \cite{alberts2023large}. BDM’nin ilaç keşfi ve gelişiminde, araştırma sürecini kolaylaştırma, maliyetleri azaltma ve çeşitli hastalıklar için yeni tedavilerin keşfedilmesine yol açmaktadır. BDM’nin devam eden gelişimi, tıp alanında inovasyonların gelişmesi için büyük bir potansiyele sahiptir.
 
BDM'lerin tıpta kullanılması, getirdiği kolaylıkların yanında çeşitli etik ve pratik zorluklar da beraberinde getirmiştir \cite{cabrera2023ethical}. Kullanıcıların duyduğu öncelikli endişelerden biri, bu modeller tarafından üretilen çıktıların doğruluğu ve güvenilirliğidir. BDM'ler tutarlı ve bağlamla ilgili yanıtlar üretebilseler de yanılmaz değildirler ve zaman zaman hatalı veya yanıltıcı bilgiler üretebilmektedirler \cite{poulain2024bias}. Bu nedenle, sağlık çalışanlarının BDM'ler tarafından sağlanan önerileri eleştirel bir şekilde değerlendirmesi ve bunları klinik uygulamalarına mantıklı bir şekilde entegre etmesi gerekmektedir. Model yanılmalarının ve yorumlanabilme sorunlarının yanında, veri gizliliği endişesi ile alınan hasta kayıt bilgilerinin korunması önem teşkil eden sorunlardandır. BDM'lerin tıp alanına entegrasyonuna özenle yaklaşılmalı, ilgili riskleri azaltarak faydalarını en üst düzeye çıkarmak için etik standartlara ve yasal düzenlemelerin gerekliliklerine uyulması sağlanmalıdır. Bu modeller gelişmeye devam ettikçe sağlık hizmetlerindeki rolleri de genişleyecek ve tıp camiasının dinamik ihtiyaçlarına uyum sağlamak için sürekli değerlendirme ve iyileştirme gerektirecektir.

TUS, Türkiye'de tıp fakültesi mezunlarının uzmanlık eğitimine başlayabilmeleri için geçmeleri gereken kritik bir sınavdır. TUS, tıbbi bilgiyi ölçen zorlu bir sınavdır. Bu sınav, adayların klinik ve temel tıp bilgilerini kapsamlı bir şekilde değerlendirmektedir. Amerika Birleşik Devletleri'nde benzer bir değerlendirme sistemi olan Amerikan Tıp Lisans Sınavı (ATLS) uygulanmaktadır. Büyük dil modellerinin tıbbi sınav sorularına verdikleri yanıtların doğruluğunu ölçmek için kullanılan medQA veriseti kullanılmaktadır. medQA, Amerikan Tıp Lisans Sınavı (ATLS) soruları üzerine yapılan çalışmalarda kullanılmıştır. Örneğin, Przystalski ve Thanki, büyük dil modellerinin ATLS sorularına verdikleri yanıtların doğruluğunu incelemiş ve bu modellerin yüksek başarı oranlarına sahip olduğunu göstermiştir \cite{przystalski2023natural}. Ayrıca ChatGPT’nin ATLS sınavındaki başarısı literatürde farklı çalışmalarda da incelenmiştir \cite{gilson2023does, mbakwe2023chatgpt}.
\color{black}

Microsoft'un 2023 yılında gerçekleştirdiği çalışma, büyük dil modellerinin tıbbi sınavlarda performansını değerlendirme konusunda önemli bir katkı sağlamıştır. Bu çalışmada, Microsoft'un geliştirdiği dil modelleri, çeşitli tıbbi sınav sorularına yanıt verme yetenekleri açısından test edilmiştir. Araştırmada, dil modellerinin doğruluk, kesinlik, duyarlılık ve F1 skoru gibi metrikler kullanılarak performansları analiz edilmiştir. Sonuçlar, Microsoft'un dil modellerinin tıbbi bilgi tabanlı sorulara yanıt verme konusunda yüksek doğruluk oranlarına sahip olduğunu ve bu modellerin tıbbi eğitim ve klinik karar destek sistemlerinde kullanılabilirliğini göstermiştir \cite{microsoft2023evaluation}.

Diğer dillerde de benzer çalışmalar mevcuttur. Çin'de, büyük dil modellerinin Çin Tıp Lisans Sınavı sorularına verdikleri yanıtlar üzerine yapılan çalışmalar, bu modellerin tıbbi bilgiye dayalı soruları yüksek doğrulukla yanıtlayabildiğini ortaya koymuştur \cite{zhang2022performance}. Bu çalışmalar, büyük dil modellerinin dil bağımsız olarak tıbbi bilgiye dayalı karar destek sistemleri olarak kullanılabileceğini göstermektedir.

Flores-Cohaila vd. \cite{flores2023performance} yaptığı çalışmada ChatGPT'nin GPT-3.5 ve GPT-4 versiyonlarının Peru Ulusal Lisans Tıp Sınavı'ndaki (PULTS) performansı değerlendirilmiştir. PULTS, 2022 veri seti kullanılarak yapılan analizde, ChatGPT'nin doğruluk oranı GPT-4 için \%86 olarak elde edilirken GPT-3.5 için ise \%77 olarak bulunmuştur. 

Wójcik vd. \cite{wojcik2023reshaping} yaptığı çalışmada ise ChatGPT-4'ün Polonya Tıp Uzmanlık Lisans Sınavı'ndaki performansı değerlendirilmiştir. Haziran 2023'te yapılan değerlendirmede, ChatGPT'nin 120 soruluk bir seti \%67.1 doğruluk oranla 80 soruya doğru yanıt vererek tamamladığı belirtilmiştir. 

Huang vd. \cite{huang2023benchmarking} tarafından yapılan çalışmada, 38.Amerikan Radyoloji Koleji Radyasyon Onkolojisi Eğitim Sınavı (ROES) ve 2022 Red Journal Gray Zone vakaları kullanılarak ChatGPT-4 modelinin radyasyon onkolojisi alanındaki performansı değerlendirilmiştir. ChatGPT-4, ROES sınavında \%74.57 doğruluk oranı ile ChatGPT-3.5'e göre daha yüksek bir performans sergilemiştir. Özellikle istatistik, merkezi sinir sistemi ve göz, pediatri, biyoloji ve fizik alanlarında güçlü olduğu; Gray Zone vakalarında ise kişiselleştirilmiş tedavi yaklaşımları sunma konusunda yüksek doğruluk ve kapsamlılık sergilediği belirtilmiştir. Ancak, klinik kullanıma hazır olmadığı ve verdiği bilgilerin doğrulanması gerektiği vurgulanmıştır.

ChatGPT'nin Tıp eğitiminde kullanılabilirliği ve ulusal tıp sınavlarındaki başarıları çeşitli araştırmalarda yer bulmuştur. Kung vd. \cite{kung2023performance} tarafından Pakistan’da yapılan bir çalışmada ChatGPT’nin öğrenim asistanlığı, kişiselleştirilmiş eğitim ve
otomatik puanlama sağlaması sebebiyle tıp eğitim ve araştırmalarda etkili bir şekilde kullanılabileceği belirtilmiştir. 

Bu çalışmalar, büyük dil modellerinin farklı dillerde tıbbi bilgi ölçme sıralamalarındaki başarımı, dil modellerinin performansı, eğitildikleri veri setlerinin kalitesine ve kapsamına bağlıdır.

OpenAI tarafından geliştirilen ChatGPT ve Google tarafından geliştirilen Gemini, geniş veri kümeleri üzerinde eğitilerek insan benzeri metin üretme ve yanıt verme yetenekleri en ileri modeller arasında yer almaktadır. Bu modeller, geniş bir bilgi tabanına sahip olmaları ve dil işleme yetenekleri sayesinde çeşitli uygulamalarda başarılı bir şekilde kullanılmaktadır. Bu bağlamda, ChatGPT ve Gemini gibi modellerin Türkçe TUS sorularına verdikleri yanıtların performansını değerlendirmek, bu modellerin Türkçe dilindeki tıbbi ölçme ve değerlendirme alanına kullanılabilirliğini anlamak açısından kritik öneme sahiptir.

\section{Yöntem}

Yapay zeka ve DDİ'deki gelişmeler, OpenAI'nin ChatGPT-4'ü, ChatGPT-4o ve Google'ın Gemini'si gibi gelişmiş modellerin üretilmesine yol açmıştır. Bu modeller, insan diline en yakın cevaplar ve metinler  üretmekte ve sorulan sorulara bir insan gibi etkileşimli şekilde cevap vermeyi amaçlamaktadır. Bu çalışmada Türkiye’de tıp öğrencilerine uygulanan TUS soruları sözü edilen üç modele  sorularak, verilen cevapların doğruluğu incelenmiştir. 
 
ChatGPT-4, OpenAI'nin Generative Pretrained Transformer (GPT), Türkçe anlamı Üretken Önceden Eğitilmiş Dönüştürücüsü'dür. Bir dönüştürücü mimarisi üzerine inşa edilmiştir ve çeşitli metin verilerinden oluşan geniş bir derlem üzerinde denetimsiz öğrenme kullanılarak eğitilmiştir. ChatGPT-4'ün temel amacı, bağlamsal olarak alakalı ve tutarlı metin yanıtları üreterek insan-bilgisayar etkileşimini geliştirmektir \cite{Openaii}. Bu model, insan benzeri metinlerin anlaşılmasının ve üretilmesinin çok önemli olduğu otomatik müşteri hizmetleri, içerik oluşturma ve eğitim yardımı gibi uygulamalarda yaygın olarak kullanılmaktadır.
 
ChatGPT-4, sadece metinsel soru-cevapların yanında, görüntü girdilerini de kabul edip bu görüntüler üzerinde de metinsel olarak cevap vermektedir. ChatGPT-4, metin ve görsel içeren belge, ekran görüntüsü ve diyagramlar olmak üzere çeşitli metin girdilerinde olduğu gibi bir başarım sergilemektedir. Bu çalışmada kullanılan TUS soruları içerisindeki 7 görsel içeren soru yer almaktadır.

13 Mayıs 2024’te duyurulan, ChatGPT-4'ün optimize edilmiş bir versiyonu olan ChatGPT-4o, çıktıların kalitesinden ödün vermeden yanıt süresi ve hesaplama verimliliği gibi başarım ölçümlerini iyileştirmek için özel olarak tasarlanmıştır. ChatGPT-4o’daki o “omni” “her şey” anlamına gelmektedir. Diğer GPT versiyonlarından farklı olarak bu versiyonda  sorgulara sadece metinsel çıktılar üretmek yerine sesli çıktılar da sunmaktadır. Geliştirilmiş optimizasyon stratejileri ile ChatGPT-4o'nun cevap verme hızı artmıştır. 
 
Bu çalışmada kullanılan üçüncü model ise Google’ın geliştirdiği Gemini'dir.  Castor ve Pollux adındaki iki sinir ağından oluşan büyük bir dil modeli olan  Gemini, çeşitli veri türlerini işlemek için denetimli ve denetimsiz öğrenme tekniklerinin bir kombinasyonunu kullanmaktadır \cite{Geminii}. Bu da onu görüntü altyazıları, video analizi ve gerçek zamanlı dil çevirisi gibi çeşitli uygulamalarda çok yönlü hale getirmektedir.

Bu çalışmanın temel araştırma sorusu, TUS özelinde hem bu modellerin kendi içinde karşılaştırılmaları hem de insan ve model başarımları arasındaki farkın incelenmesidir.  Türkiye de uygulanan TUS 2021 1. Döneme ait 120’şer sorudan oluşan Klinik Tıp Bilimleri Testi ve Temel Tıp Bilimleri Testi bu çalışmada kullanılan bu üç modelle -ChatGPT-4, ChatGPT-4o ve Google Gemini- tek tek sorgulanarak cevapları Ölçme Seçme ve Yerleştirme Merkezi’nin (ÖSYM) vermiş olduğu cevap anahtarı ile karşılaştırılmıştır. 

Üç modele de veri kümesindeki sorular teker teker yazılı biçimde sorulmuştur. Bütün deneylerde yanlış verilen cevaplara "Emin misin?" şeklinde tekrar sorulmuştur. ChatGPT-4 ve ChatGPT-4o "Emin misin?" sorusuna verdiği ikinci yanıtta değişiklik yapmamıştır. Öte yandan, Google Gemini, ChatGPT versiyonlarından farklı olarak verdiği ilk cevaptan farklı cevaplar üretmiştir. Bu durumda, ikinci cevabı asıl olarak kayıt edilmiştir. 

\subsection{Veri Kümesi}
Bu çalışmada ÖSYM'nin yapmış olduğu 2021 TUS 1. Dönem sınavı temel veri kümesi olarak kullanılmıştır.  Bu sınav iki bölümden oluşmaktadır: Klinik Tıp Bilimleri Testi (KTBT) ve Temel Tıp Bilimleri Testi (TTBT). Her bir bölümde 5 seçenekli 120 adet soru bulunmaktadır. Sınavda 7 adet görsel içerikli sorular bulunmaktadır. Bunlardan; 2 tanesi KTBT sınavına ait, 5 tanesi TTBT sınavına ait görsel içerikli sorulardır. Bu görseller 3 ilüstrasyon, 2 grafik, 1 şekil ve 1 fotoğraftan oluşmaktadır. Sınav kitapçığı ve cevap anahtarı ÖSYM'nin web sitesinde kamuya açık olarak PDF dosya formatında yayınlanmıştır \cite{tustemel, tusklinik}. Bu dosya elle temizlenerek soruların metinleri ayıklanmıştır. Sorulardaki görseller de benzer şekilde elle ayıklanmış veya ekran görüntüsü alınıp PNG formatında kaydedilmiştir. Sorulara \url{https://www.osym.gov.tr/TR,15072/tus-cikmis-sorular.html} linkinden ulaşılabilmektedir.

TUS sınavında metin tabanlı sorular ile adayların bilgileri ölçülmekte ve tanı koyma becerileri değerlendirmektedir. Bu sınavda metinsel içeriğe ek olarak, grafik barıdıran sorular da yer almaktadır. Verilen grafik ile adayların yorumlama yeteneklerinin ölçülmesi amaçlanmaktadır.

Literatürdeki diğer çalışmalarda yer verilen test sınavlarından farklı olarak TUS soruları 5 şıktan oluşmaktadır. Diğer çalışmalardaki \cite{perrault2024artificial} 4 şıklı sorularda doğru cevabı tahmin etme olasılığı \%25 iken 5 şıklı soru için \%20'dir, bu da rasgele tahminin daha az etkili olmasını sağlamaktadır. Ayrıca seçenek sayısı daha fazla olduğu için doğru cevabı bulmak daha fazla düşünme gerektirebilir ki bu durum, modelin daha fazla bilgi işlemesi ve analiz etmesini gerektirip cevap verme süresini uzatabilir. 

\section{Deney Sonuçları ve Tartışma}
Bu bölümde, çalışmada kullanılan veri kümeleri detaylı bir şekilde açıklanmıştır. Ayrıca her bir deney kurgusu ve deneylerden elde edilen sonuçlar gösterilmiştir. Her bir modelin başarımı detaylı olarak ele alınmış ve  sonuçlar karşılaştırmalı olarak sunulmuştur. 

Bu çalışmada, görsel içerikli soruların tümü için ChatGPT-4 ve ChatGPT-4o bir cevap vermiştir. TTBT’de yer alan 2 numaralı soru bir MR görüntüsü içerirken KTBT’deki 54 numaralı soru bir çeşit deri hastalığına sahip bir çocuğun görüntüsünü içermektedir. Gemini, TTBT’deki 2 numaralı soru ve KTBT’deki 54 numaralı soru olmak üzere bu sorulardan ikisinin görüntüsünü işleyemeyip bir cevap vermemiştir.  Bu durumla ilgili herhangi bir geri bildirimde bulunmamıştır. Gemini bu noktada diğer modellere göre geride kalmıştır.

\begin{figure}[!ht]
	\centering\hspace*{-0.80mm}
	\subfloat[]{\includegraphics[width=0.45\textwidth]{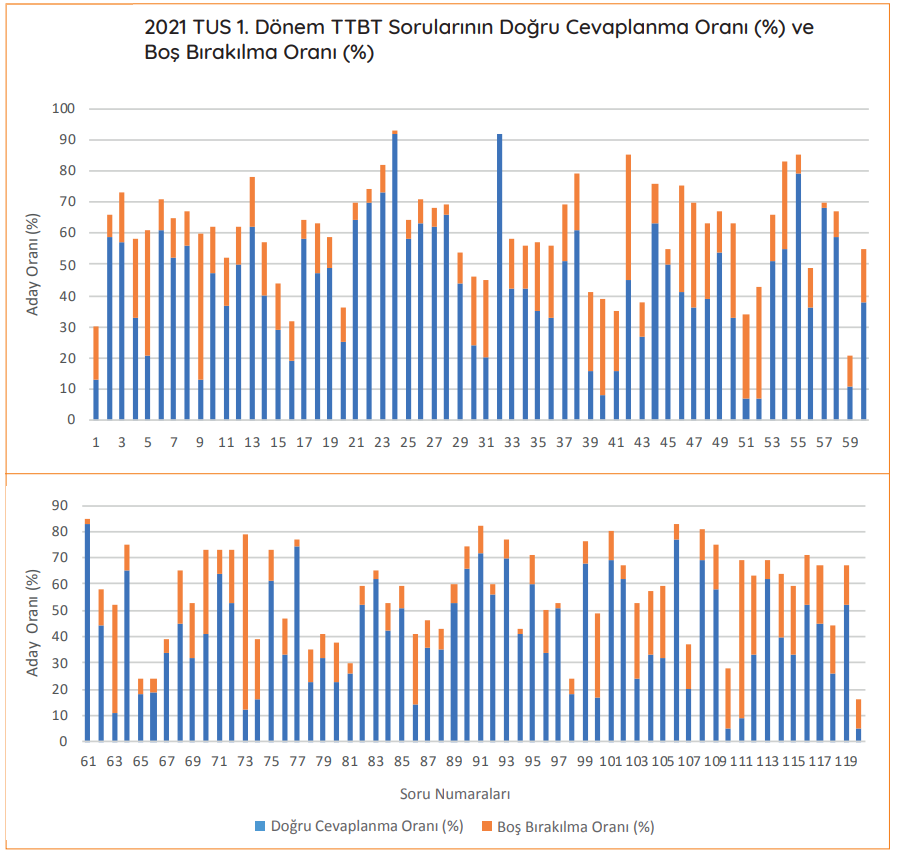}
		\label{ttbt}}
		\hfil
	\subfloat[]{\includegraphics[width=0.45\textwidth]{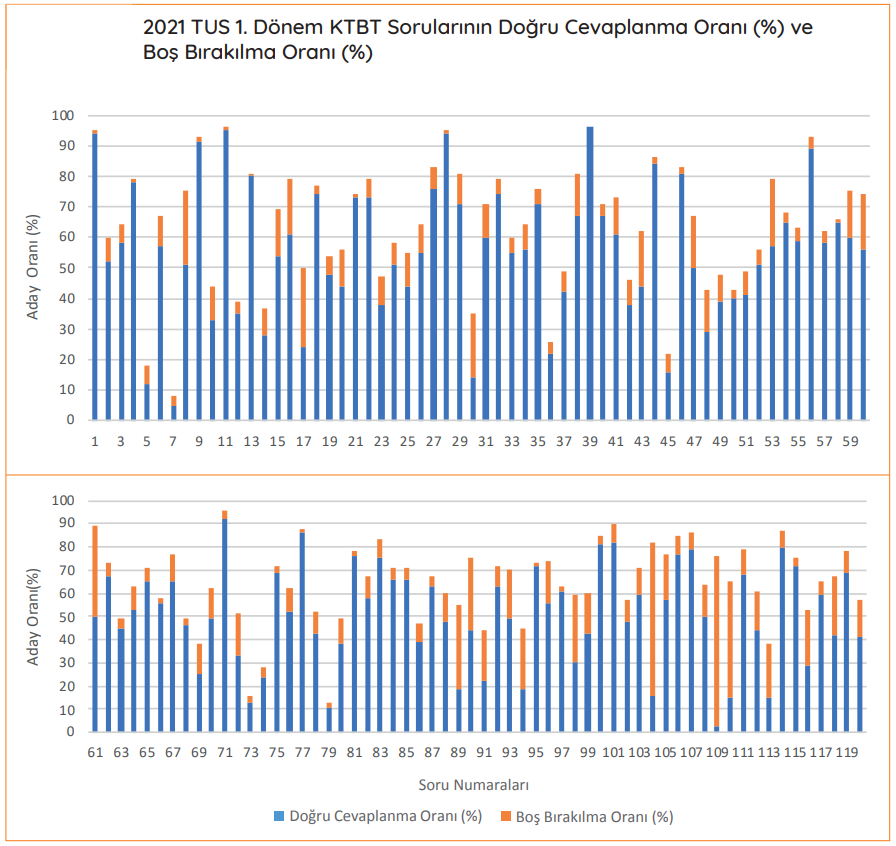}
		\label{ktbt}}
	\hfil
        \caption{2021 TUS 1. Dönem TTBT ve KTBT sorularının doğru cevaplanma oranı(\%) ve boş bırakılma oranı (\%)}
	\label{figure2}
\end{figure}

ÖSYM’nin hazırlamış olduğu değerlendirme raporunda \cite{tusdegerlendirmeraporu} 2021 TUS 1. Dönemde TTBT için 18436, KTBT içinse 18280 adayın sınavı geçerli sayılmıştır. TTBT doğru cevap sayısı ortalaması 51.63 ve KTBT doğru cevap sayısı ortalaması 63.95 olarak hesaplandığı raporlanmıştır. Bu doğrultuda, KTBT doğru cevap sayısı ortalaması TBTT doğru cevap sayısı ortalamasından daha yüksek olduğu görülmektedir. Bununla birlikte, ChatGPT-4o ise her iki testte de başarı olarak ortalamanın üzerindedir. TTBT ve KTBT’de bulunan 120 sorunun tamamını doğru cevaplayan aday bulunmamakta olup TTBT’de 1 adayın 106 soruyu; KTBT’de ise 1 adayın 113 soruyu doğru cevaplayarak en yüksek doğru cevap sayısına ulaştığı raporlanmıştır.Ayrıca TTBT’ye giren adayların doğru cevap sayılarının 32-58 aralığında; KTBT’ye giren adayların ise doğru cevap sayıları 45-75 aralığında yığıldığı belirtilmiştir.

Şekil \ref{ttbt}'da da görüldüğü gibi TTBT’de 24 ve 32 numaralı soruların \%92 ile en yüksek, 110 ve 120 numaralı soruların ise \%5 ile en düşük doğru cevaplanma oranına sahip olduğu belirtilmiştir. ChatGPT-4o ve Gemini 24, 32 ve 110; ChatGPT-4 ise 120 numaralı sorular için cevabı doğru bir şekilde vermiştir. Bununla birlikte, \%67 oranla 73 numaralı soru en çok boş bırakılan soru olup TTBT sorularının boş bırakılma oranlarının ortalaması \%16 olarak hesaplanmıştır. Bu kapsamda, üç model de en çok boş bırakılan soruyu cevap anahtarına uygun bir şekilde cevaplamıştır.

Şekil \ref{ktbt}'deki KTBT’de ise 39 numaralı soru \%96 ile en yüksek, 109 numaralı soru \%3 ile en düşük doğru cevaplanma oranı olarak belirlenmiştir. Benzer şekilde, üç model de 39 numaralı soruyu cevap anahtarına uygun olarak doğru bir şekilde cevaplamıştır. 109 numaralı soru, adayların \%73’ü tarafından boş bırakılmış olup bu sorunun adaylar için zorlayıcı bulunmuş olduğu belirtilmiştir. ChatGPT-4o ise bu soru için yine doğru cevabı sunmuştur. TTBT’de olduğu gibi KTBT’nin boş bırakılma oranlarının ortalaması \%11.1 olarak hesaplandığı belirtilmiştir. 

\begin{figure}[!ht]
	\centering\hspace*{-0.80mm}
	\subfloat[]{\includegraphics[width=0.45\textwidth]{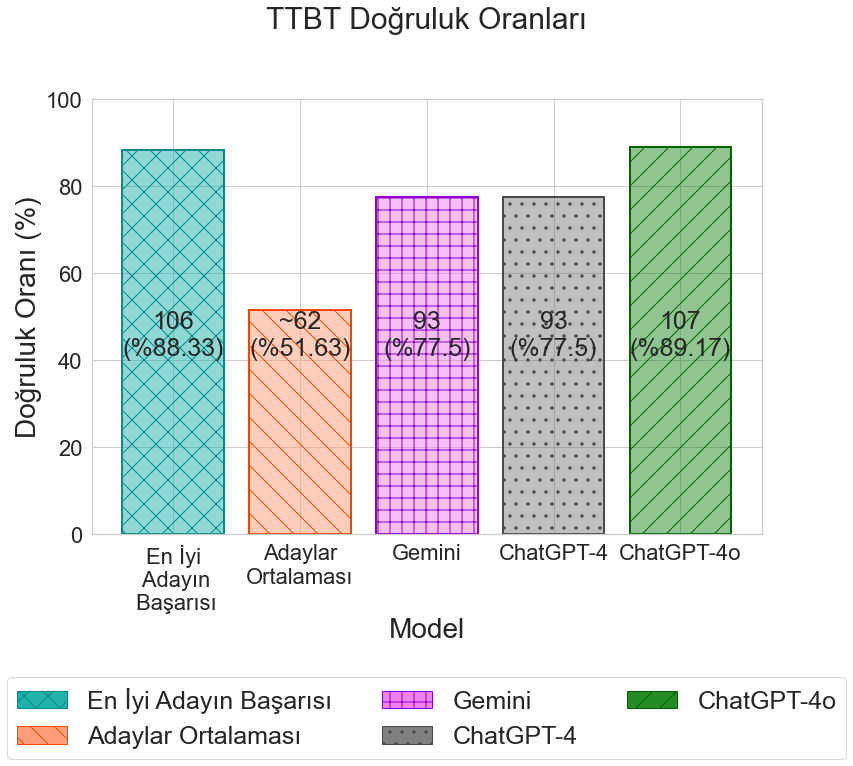}
		\label{2ttbt}}
		\hfil
	\subfloat[]{\includegraphics[width=0.45\textwidth]{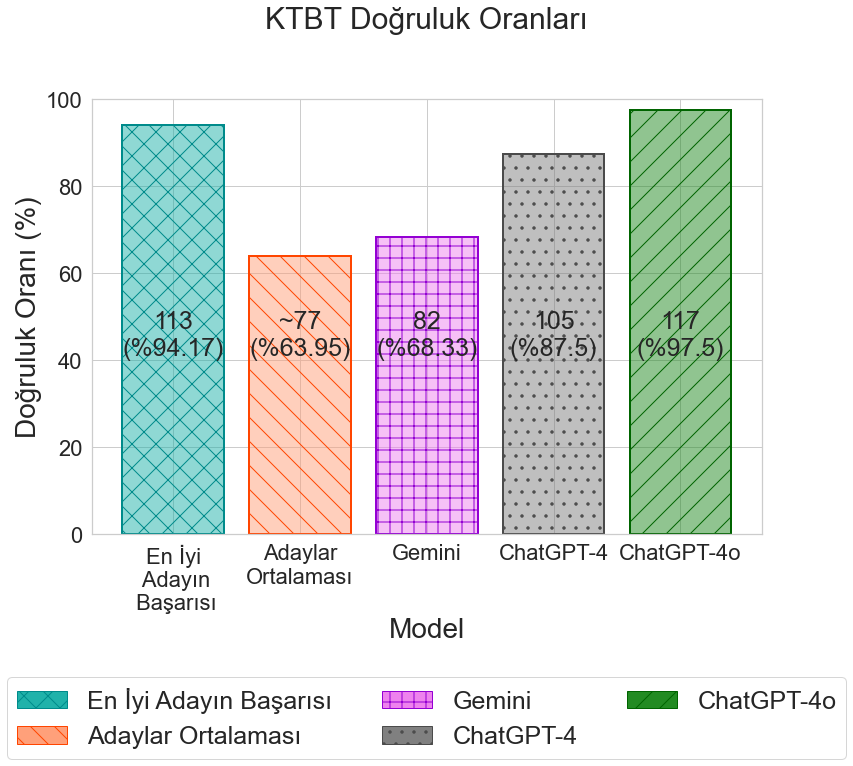}
		\label{2ktbt}}
	\hfil
        \caption{Modellerin TTBT ve KTBT'deki doğruluk oranları}
	\label{figure4}
\end{figure}

Tüm cevaplar göz önüne alındığında TTBT ve KTBT testlerinde yapay zeka modelleri ortalama bir insandan çok daha iyi performans sergilemiştir. Şekil \ref{2ttbt} ve Şekil \ref{2ktbt}'de özellikle ChatGPT-4o modeli, doğru cevap sayıları ile öne çıkmaktadır. ChatGPT-4o’dan alınan cevaplar ile cevap anahtarı olarak verilen sonuçlar karşılaştırıldığında ChatGPT-4o, TTBT ve KTBT testlerinde sırasıyla 107 ve 117 soruyu doğru cevaplamıştır.

Şekil \ref{2ttbt} ve Şekil \ref{2ktbt}'de yer verilen grafikler incelendiğinde 2021 TUS 1. Dönemi değerlendirmesi kapsamında, TTBT ve KTBT testlerinde yapay zeka modellerinin performansına dair yapılan analizler sonucunda TTBT'de Gemini ve ChatGPT-4 modellerinin doğruluk oranları \%77.5 iken ChatGPT-4o modelinin doğruluk oranı \%89.17 olarak elde edilmiştir. Bu durum, ChatGPT-4o modelinin en yüksek doğruluk oranına sahip olduğunu göstermektedir. En iyi adayın 120 soruda 106 soruyu doğru yapması ile başarım oranının \%88.33 olduğu söylenebilir. Adaylar ortalaması dikkate alındığında ise adayların ortalama olarak 62 soru doğru yaptığı belirtilebilir.

KTBT'de  ChatGPT-4 modeli \%87.5 oranında doğru sonuç vermiştir. ChatGPT-4o modeli ise \%97.5 doğruluk oranı ile yine en yüksek doğruluğa sahiptir. Şekil \ref{2ktbt} incelendiğinde KTBT’de Gemini modelinin doğruluk oranı \%68.33 olarak belirlenmiş olup, ChatGPT-4 ve ChatGPT-4o modellerine kıyasla daha düşük bir performansa sahip olduğu söylenebilir. TTBT'dekine benzer şekilde, en iyi adayın 120 soruda 113 soruyu doğru yapması ile başarım oranının \%94.17 olduğu, adaylar ortalaması dikkate alındığında ise adayların ortalama olarak 77 soru doğru yaptığı belirtilebilir.

\section{Sonuç}

Bu çalışmada, ChatGPT-4, ChatGPT-4o ve Gemini dil modellerinin 2021 TUS 1. Dönem TTBT ve KTBT sınav sorularını cevaplama başarımları değerlendirilmiştir. Yapılan deneyler, bu yapay zeka modellerinin, ortalama bir insanın performansından çok daha yüksek doğruluk oranlarına sahip olduğunu ortaya koymuştur. Özellikle ChatGPT-4o modeli, her iki test grubunda da en yüksek doğruluk oranına ulaşarak, en yüksek başarıma sahip olduğu görülmüştür. Elde edilen sonuçlar, kullanılan dil modellerinin en iyi adaydan daha iyi performans gösterdiğini açıkça ortaya koymuştur. Bu bulgular ışığında, araştırma sorusunun deneysel olarak yanıtlandığını ve hipotezin doğrulandığı görülmektedir.

Modellerin doğru cevap oranlarının yanı sıra, ChatGPT-4o modelinin diğer modellere kıyasla anlamlı derecede daha yüksek doğruluk oranlarına sahip olduğunu gözlemlenmiştir. ChatGPT-4 ve Gemini modelleri de genel olarak iyi performans sergilemiştir. Bununla birlikte, özellikle görsel içerikli sorularda Gemini modelinin bazı zorluklar yaşadığı gözlemlenmiştir. ChatGPT-4 modeli, birçok soruya doğru cevap vermiş ancak ChatGPT-4o modeli kadar yüksek doğruluk oranlarına ulaşamamıştır. 

Tüm bu bulgular, yapay zeka destekli dil modellerinin tıbbi çeşitli alanlarda etkili bir şekilde kullanılabileceğini ve bu alanda önemli bir potansiyele sahip olduğunu göstermektedir. Özellikle ChatGPT-4o modelinin yüksek doğruluk oranları, yapay zekanın tıbbi soruları yanıtlamada ve tıbbi karar destek sistemlerinde kullanılabileceği şeklinde değerlendirilmiştir. 

\bibliographystyle{unsrt}  
\bibliography{references}

\end{document}